\documentclass[smallextended]{llncs}
\usepackage[latin1]{inputenc}
\usepackage{amsmath}
\usepackage{amsfonts}
\usepackage{stmaryrd}
\usepackage{amssymb}
\usepackage{color}
\usepackage{mathrsfs}
\usepackage{algorithmic}
\usepackage{algorithm}
\usepackage{array}
\usepackage{epsfig, subfigure, sidecap}
\usepackage{hyperref}
\definecolor{darkGreen}{rgb}{0.01,0.5,0.01}

\newcommand{\ioplb}{c}
\newcommand{\kij}{{k i j}}

\author{ }
\institute{ }
\title{A Perfectly Invertible Rank Order Coder}
\begin{document}
\maketitle
\begin{abstract}
Our goal is to revisit rank order coding by proposing an original exact decoding procedure for it. Rank order coding was proposed by Simon Thorpe et al. to explain the impressive
performance of our visual system in recognizing objects. It is based on the hypothesis that the retina represents the visual stimulus by the order in which its cells are activated. A rank order coder/decoder was then designed involving three stages~\cite{VanRullen01}: (i) A model of the stimulus transform in the retina consisting in a redundant filter bank analysis; (ii) A sorting stage of the filters according to their activation degree; (iii) A straightforward decoding procedure consisting in a weighted sum of the most activated filters. 
Focusing on this last stage, it appears that the decoding procedure employed yields reconstruction errors that limit the model Rate/Quality performances when used as an image codec. The attempts made in the literature to overcome this issue are time consuming and alter the coding procedure, or are lacking mathematical support and are infeasible for standard size images. Here we solve this problem in an original fashion by using the frames theory, where a frame of a vector space designates an extension for the notion of basis. First, we prove that the analyzing filter bank considered is a frame, and then we define the corresponding dual frame that is necessary for the exact image reconstruction. Second, to deal with the problem of memory overhead, we design a recursive Out-Of-Core blockwise algorithm for the computation of this dual frame. 
Our work provides a mathematical formalism for the retinal model under study and specifies a simple and exact reverse transform for it. Furthermore, the framework presented here can be extended to several models of the visual cortical areas using redundant representations.
\end{abstract}

\parskip 0.175pt
\section{Introduction}
\label{sec:intro}

Neurophysiologists made substantial progress  in better understanding the early processing of the visual stimuli. Especially, several efforts proved the ability of the retina to code and transmit  a huge amount of data under strong time and bandwidth constraints~\cite{Thorpe90,Meister99}. 
The retina does this by means of a neural code consisting in a set of electrical impulses: the spikes. 
Based on these results, our aim is to use the computational neuroscience models that mimic the retina behavior to design novel lossy coders for static images. 

 Although there is no clear evidence on how the spikes encode for the visual stimuli, we assume in this paper that the relevant encoding feature is the order in which the retina cells emit their first spike given the stimulus onset. This choice was motivated by Thorpe et al. neurophysiologic results on ultra-rapid stimulus categorization~\cite{Thorpe90,Marlot96}. Authors showed that still image classification can be achieved by the visual cortex within very short latencies of about 150 ms or even faster. As an explanation, it was stated that: 
\textit{There is information in the order in which the cells fire}, and thus the temporal ordering can be used as a code.  
This code termed as rank order coding (ROC)  was further studied 
and yielded the conception of a classical retina model~\cite{VanRullen01}. 

However, one major limitation of the ROC coder defined in~\cite{VanRullen01} prevents us from its use for the design of image codecs. It is the inaccuracy of the proposed decoding procedure. Indeed, the bio-inspired retina model that generates the spikes is based on a redundant filter bank image analysis, where the considered filters are not strictly orthogonal. Thus, the filter overlap yields reconstruction errors that limit the Rate/Quality performance~\cite{Perrinet04,Bhattacharya2010,Masmoudi2010}. 
Efforts to correct this issue followed two main approaches. A first one consisted in inverting directly the transform operator to obtain a reverse filter bank as in~\cite{Sen07,Bhattacharya2010}. The method presented is based on a pseudo-inversion. Though interesting, we will prove that this method lacks mathematical support.  
Besides the procedure used deals with a high dimension matrix and thus is infeasible, as such, for standard size images.
A second approach relies on Matching Pursuit (MP) algorithms as in~\cite{Perrinet04,Bhattacharya2010}. These methods are time consuming and alter the coding procedure. In addition, the MP approach depends on the order in which the ``match and update'' mechanism is performed, and this makes the coding procedure depend on the stimulus itself.  

In this paper, we give an original solution relying on the mathematical concept of frames and dual frames, which is an extension of the notion of basis~\cite{
Kovacevic2008}. First, we express the model in a matrix-based fashion. Then we prove that the analyzing filter bank as we define it is a frame. Finally, we construct the corresponding dual frame that is necessary for the exact image reconstruction. Besides we design an out-of-core algorithm for the computation of the dual frame which resolves the encountered issues of memory overhead. 
Our method  requires the computation of a reverse operator once for all and keeps the bio-inspired coding scheme unchanged. Thanks to this approach, we show that the reconstructed image that we get is equal to the original stimulus. 

This paper is organized as follows: 
In Section~\ref{sec:imageTransformForRoc}, we present the three stages of the rank order coding/decoding method. 
Then in Section~\ref{sec:decodageExact} we 
define an exact decoding scheme through the construction of a dual frame. Finally in Section~\ref{sec:results}, we summarize the results and show the gain that we obtain in terms of Rate/Quality tradeoff.

\section{The rank order coder/decoder: Three stages}
\label{sec:imageTransformForRoc}
\subsection{The image transform: A bio-inspired retina model}
\label{sec:imageTransform}
\begin{figure}[h!]
\centerline{
\subfigure[\label{fig:cameraman}]{\includegraphics[width=0.3\textwidth]{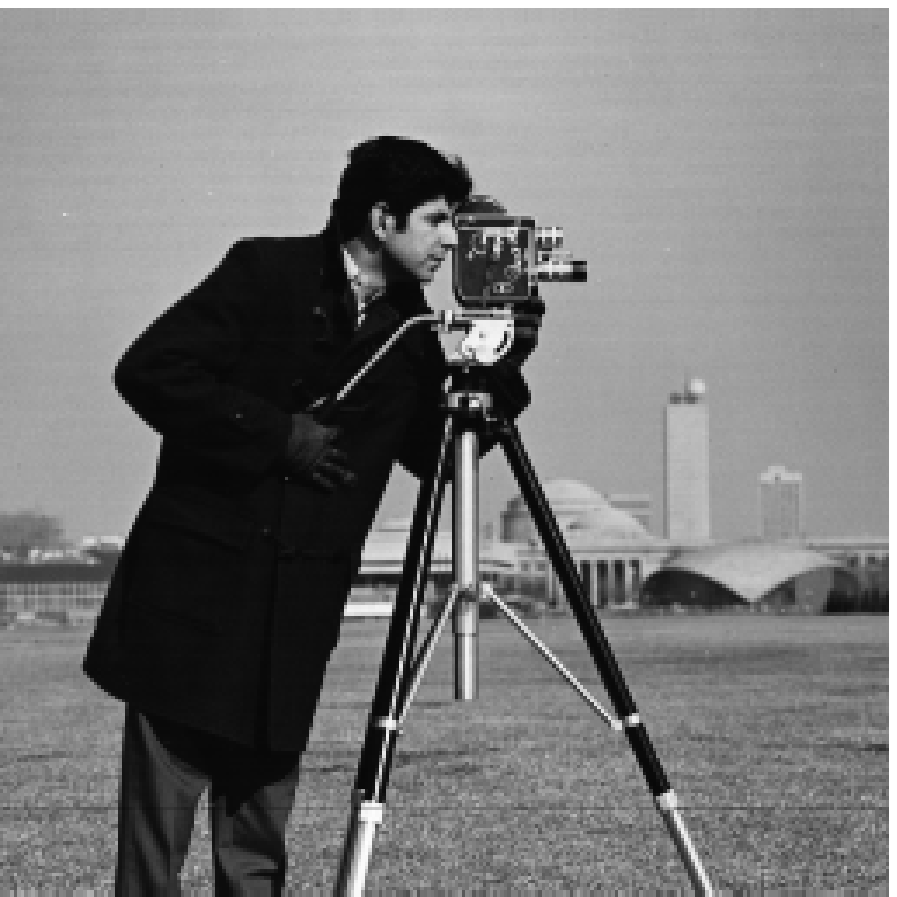} }
\subfigure[\label{fig:dyadicGrid}]{\includegraphics[height=0.21\textwidth]{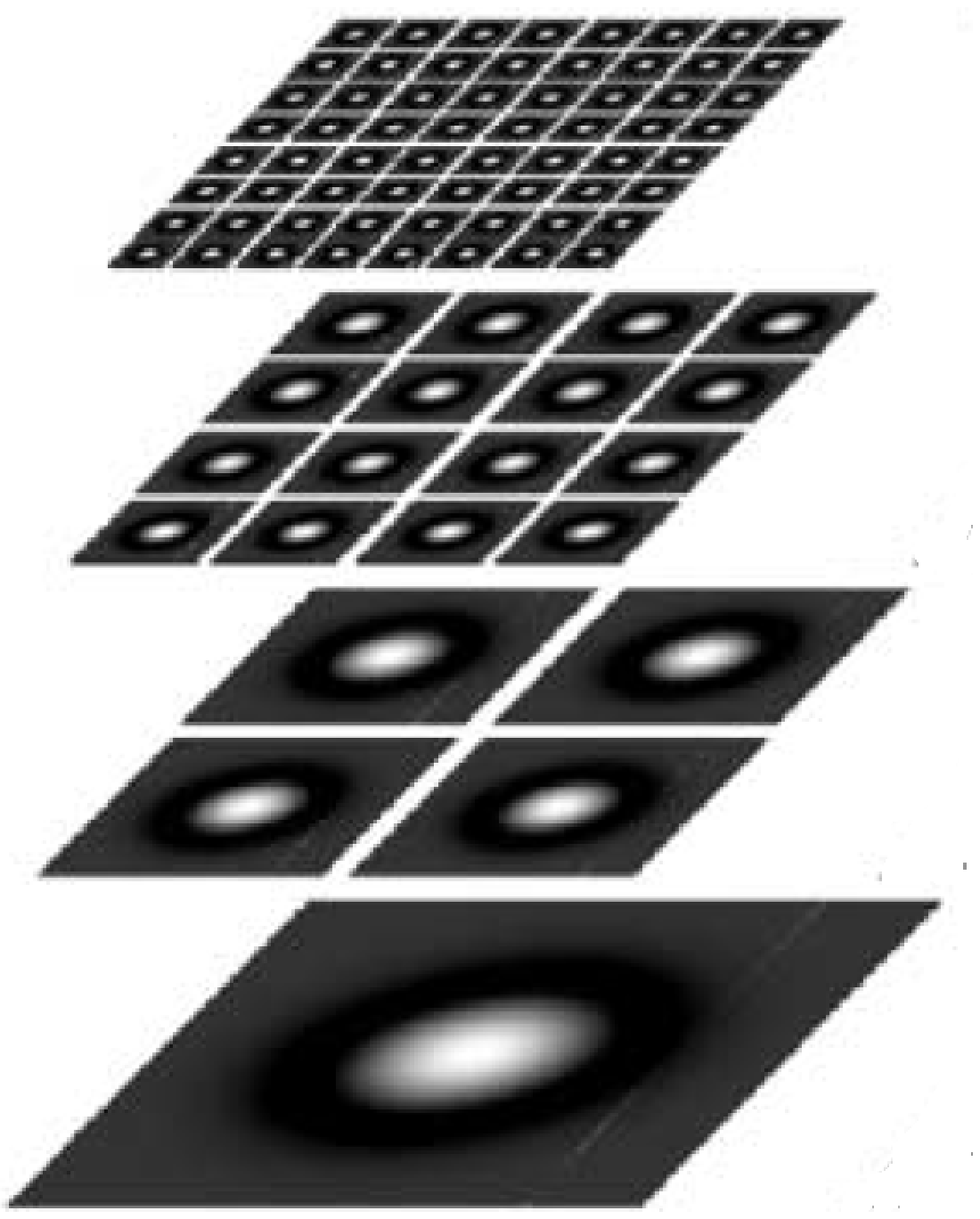}}
\subfigure[\label{fig:transformeeCameraman}]{\includegraphics[height=0.3\textwidth]{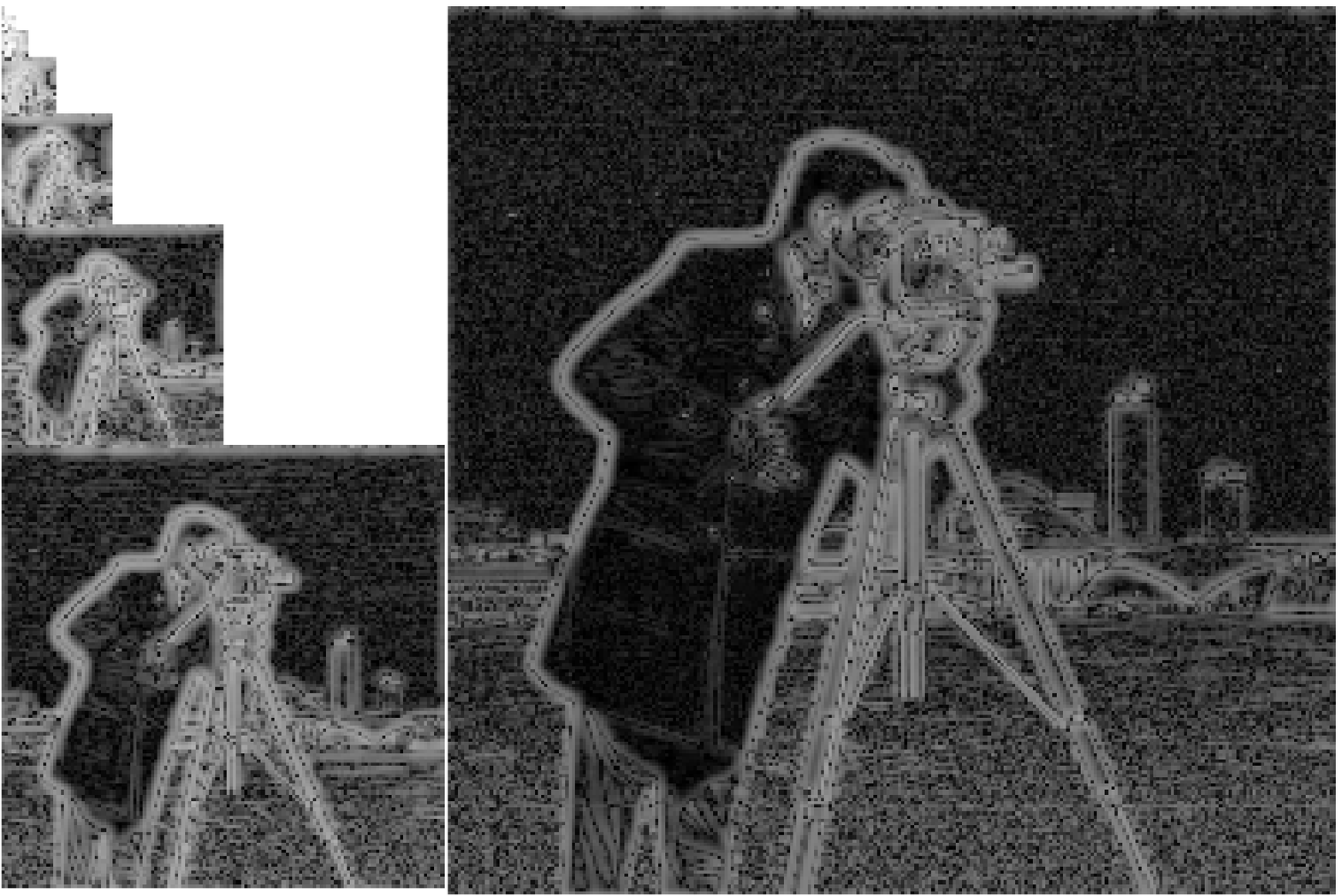} }
}
\caption{\label{fig:coding}Illustration of the retinal image transform applied on {\itshape cameraman}~\ref{fig:cameraman}. Image size is $257{\times}257$ pixels. 
~\ref{fig:dyadicGrid}: Example of a dyadic grid of DoG's used for the image analysis (from~\cite{VanRullen01}).
~\ref{fig:transformeeCameraman}: The transform result showing the  subbands.
}
\end{figure}

Neurophysiologic experiments have shown that, as for classical image coders, the retina encodes the stimulus representation in a transform domain. The retinal stimulus transform is performed in the cells of the outer layers. Quantitative studies 
have proven that the outer cells processing can be approximated by a linear filtering. In particular, the authors in~\cite{Field94} proposed the largely adopted DoG filter which is a weighted difference of spatial Gaussians that is defined as 
$
DoG (x, y) = w^c G_{{\sigma}^c}(x,y)-w^s G_{{\sigma}^s}(x,y),
$ 
where $w^c$ and $w^s$ are the respective positive weights of the center and surround components of the receptive fields, 
 $\sigma^{c}$ and $\sigma^{s}$ are the standard deviations of the Gaussian kernels $G_{{\sigma}^c}$ and $G_{{\sigma}^s}$, such that $\sigma^{c}<\sigma^{s}$.
The DoG cells can be arranged in a dyadic grid $\Gamma$ of $K$ layers to sweep all the stimulus spectrum as shown in Figure~\ref{fig:dyadicGrid}~\cite{VanRullen01,Perrinet04,Masmoudi2010}.  Each layer $0\leqslant k < K$ in the grid, is tiled with filtering cells having a scale $k$ 
and generating a transform subband ${B}_k$, that we denote by $DoG_k$:
\begin{equation}
\label{eq:dogk}
DoG_k (x, y) = w^{c} G_{{\sigma}^c_k}(x,y)-w^s G_{{\sigma}^s_k}(x,y),
\end{equation} 
where $\sigma^{c}_{k+1}=\frac{1}{2}\sigma^{c}_{k}$ and $\sigma^{s}_{k+1}=\frac{1}{2}\sigma^{s}_{k}$. Each $DoG_k$  filter has a size of \linebreak $(2M_k+1) \times (2M_k+1)$, with $M_k=3\,\sigma_k^s$. Authors in~\cite{VanRullen01} chose the biologically plausible parameters $w^{c}=w^{s}=1$, ${\sigma}^c_k = \frac{1}{3}{\sigma}^s_k \; \forall k$, and ${\sigma}^c_{K-1} = 0.5$ pixel. 

In order to measure the degree of activation $\ioplb_\kij$ of a given retina cell, such that $(k,i,j)\in\Gamma$, we compute the convolution of the original image $f$ by the $DoG_k$ filter. Yet each layer $k$ in the dyadic grid $\Gamma$ is undersampled with a step of $2^{K-k-1}$ pixels with an original offset of $\lfloor 2^{K-k-2} \rfloor$ pixels, where $\lfloor . \rfloor$ is the floor operator. 
Having this, we define the function $u_k$, such that the $\ioplb_{\kij}$ coefficients are computed at the locations $\big(u_k(i), u_k(j)\big)$ as follows:
\begin{equation}
\label{eq:uk}
u_k(i) = \lfloor 2^{K-k-2} \rfloor + 2^{K-k-1}i, \forall k\in \llbracket 0, K-1 \rrbracket.
\end{equation}
$u_k$ can be seen as an undersampling function. We notice that $u_{K-1}(i)=i$, and that the other functions ${(u_k)}_{k\in\llbracket 0, K-2 \rrbracket}$ are undersampled versions of $u_{K-1}$. 
$c_\kij$ is then computed as follows:
\begin{equation}
\label{eq:cijs}
\ioplb_\kij = \sum_{x=u_k(i)-M_k}^{u_k(i)+M_k}\sum_{y=u_k(j)-M_k}^{u_k(j)+M_k}DoG_{k}(u_k(i)-x, u_k(j)-y) \, f(x, y).
\end{equation}
The transform specified generates a vector $c$ of approximatively $\frac{4}{3} N^2$ $\ioplb_\kij$ coefficients for an $N^2$-sized image, as the architecture of this transform is similar to a Laplacian pyramid~\cite{Burt1983}. An example of such a transform is shown in Figure~\ref{fig:coding}.
%
\subsection{Sorting: The generation of the rank order code}
\label{sec:roc}
Thorpe et al~\cite{Thorpe90,Marlot96} proposed that the order in which spikes are emitted encodes for the stimulus. This yielded the ROC which relies on the following simplifying assumptions:
(i) From stimulus onset, only the first spike emitted is considered; 
(ii) The time to fire of each cell is proportional to its degree of excitation;
(iii) Only the order of firing of the neurons encodes for the stimulus.\\
Such a code gives a biologically plausible interpretation to the rapidity of the visual stimuli processing in the human visual system. Indeed, it seems that most of the processing is based on feed-forward mechanism before any feedback occurs~\cite{Marlot96}.
So, the neurons responses $(\ioplb_\kij)_{\kij\in \Gamma}$ defined in Equation~\eqref{eq:cijs} are sorted in the decreasing order of their amplitude $|\ioplb_{\kij}|$. 

The final output of this stage, the ROC, is then a sorted list of $N_s$ couples $\left(p, \ioplb_{p}\right)$ such that $\vert\ioplb_{p}\vert\geqslant\vert\ioplb_{p+1}\vert$, with $p$ being the index of the cell defined by: $p(k,i,j)=k\,N_k^2+i\,N_k+j$ and $N_k^2$ the number of cells in the subband $B_k$.
Here, the generated series $\left(p, \ioplb_{p}\right)_{0 \leqslant p < N_s}$ is the only data transmitted to the decoder.
\subsection{Decoding procedure of the rank order code} 
\label{sec:decodageRoc}
We consider the set of the first $N_s$ highest cell responses forming the ROC of a given image $f$. In~\cite{VanRullen01}, the authors defined ${\tilde{f}}_{N_s}$, the estimation of $f$ by:
\begin{equation}
\label{eq:definitionReconstruction}
{\tilde{f}}_{N_s}(x, y) = 	\sum_{p(k,i,j) = 0 }^{N_s-1}
					 {\ioplb}_p \, DoG_k(u_k(i)-x, u_k(j)-y).
\end{equation}
Equation~\eqref{eq:definitionReconstruction} gives a  progressive reconstruction depending on $N_s$. Indeed, one can restrict the code to the most valuable coefficients $\ioplb_p$, i.e the most activated cells of the retina. This feature makes the coder be scalable~\cite{Masmoudi2010}.

\begin{figure}[h!]
\centerline{
\subfigure[\label{fig:cameramanTilde}]{\includegraphics[height=0.3\textwidth]{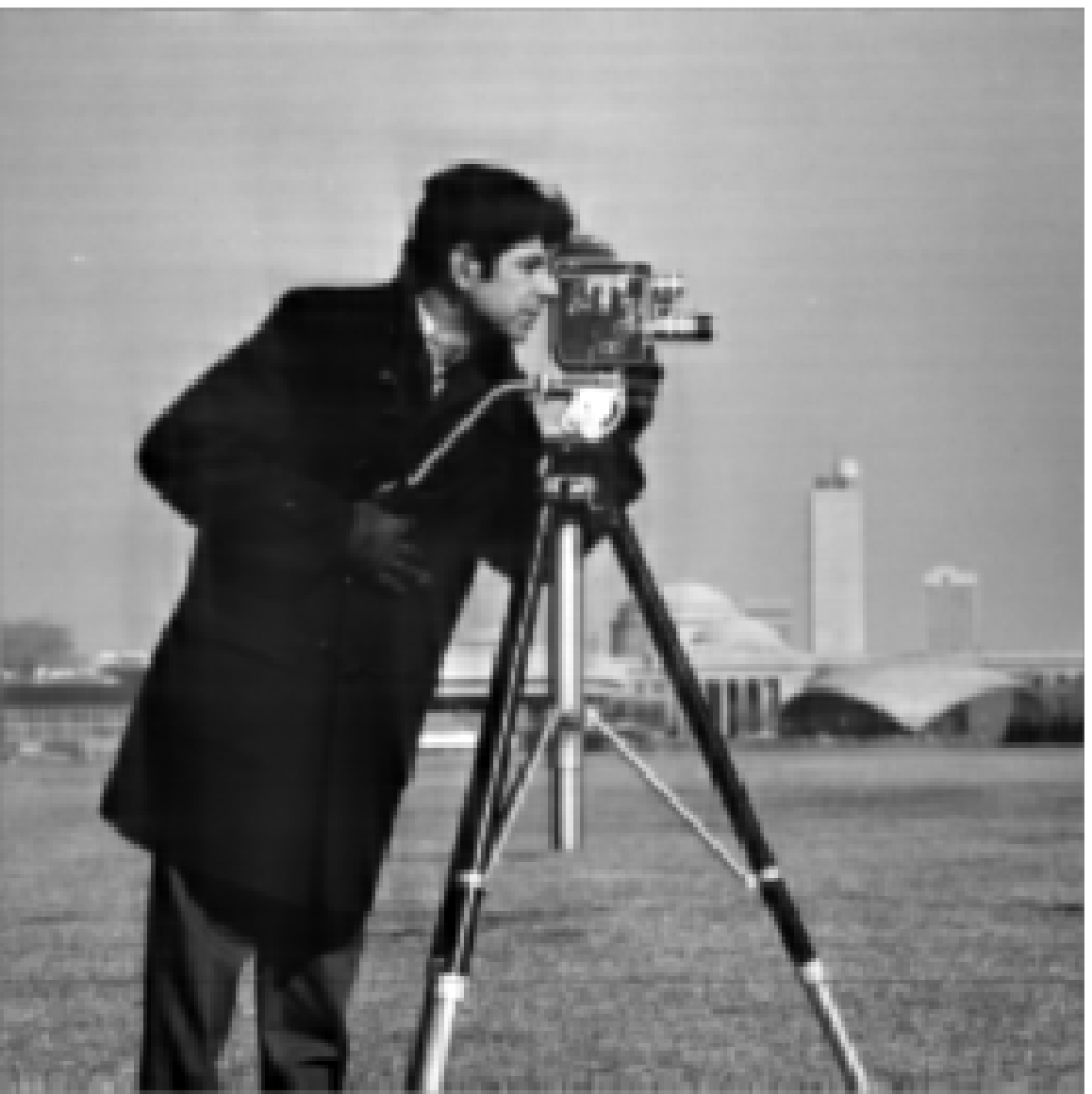} }
\subfigure[\label{fig:cameramanTildeErreur}]{\includegraphics[height=0.3\textwidth]{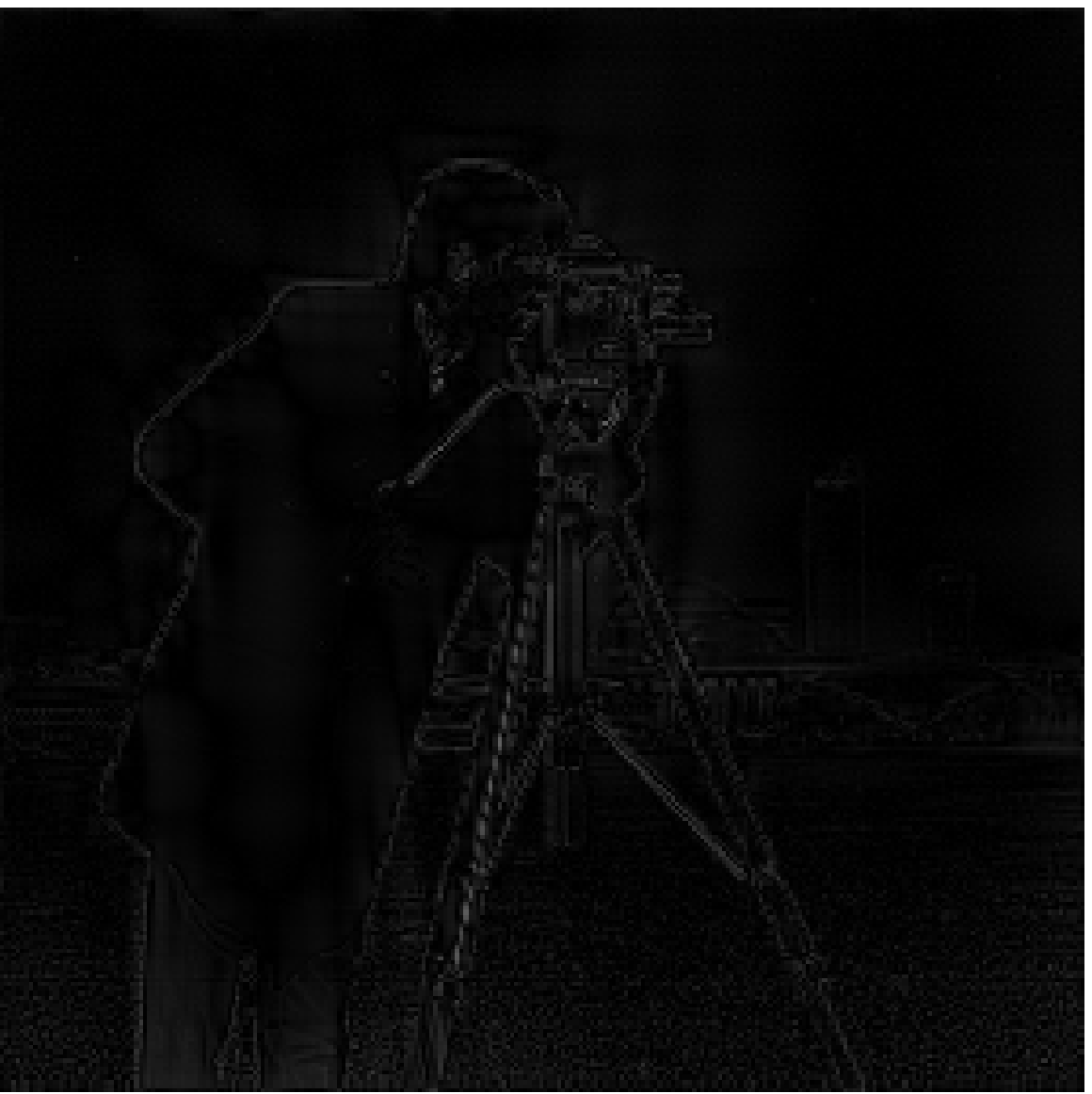} }  
\subfigure[\label{fig:cameramanTildeErreurLog}]{\includegraphics[height=0.3\textwidth]{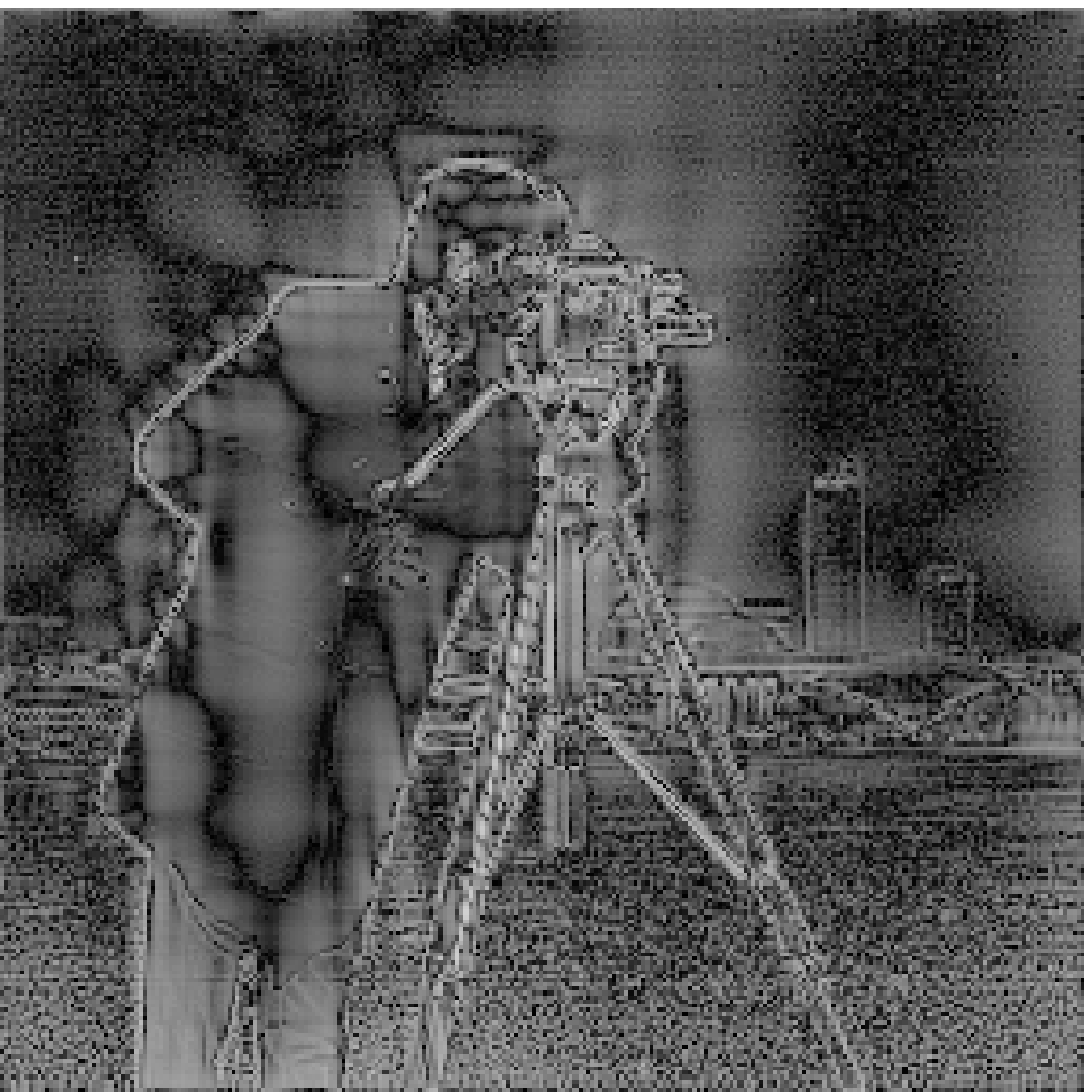} }
}
\caption{\label{fig:decoding-original}Result of the decoding procedure with the original approach using all of the retina cells responses. 
~\ref{fig:cameramanTilde}: Reconstructed image. The PSNR quality measure of $\tilde{f}_{N_s}$ yields 27 dB.
~\ref{fig:cameramanTildeErreur}: Error image: high frequencies are the ones that are the most affected by this approach.
~\ref{fig:cameramanTildeErreurLog}: Same as (b) but in a logarithmic scale.
}
\end{figure}

An example of such a reconstruction is given in Figure~\ref{fig:decoding-original}, with all the retina cells taken into account. Figure~\ref{fig:decoding-original} also shows that the retina model decoding procedure, though giving a good approximation of the stimulus, is still inaccurate. In this example, reconstruction quality is evaluated to $27$ dB of PSNR.

\section{Inverting the bio-inspired retina model}
\label{sec:decodageExact}

\subsection{Introduction of a low-pass scaling function}
\label{sec:scalingFunction}
We introduce a low-pass scaling function in the filter bank used for image analysis. This modification do not alter the ROC coder architecture and has both a mathematical and a biological justification.

\begin{figure}
\centerline{
\subfigure[\label{fig:spectreSemiLogY}]{\includegraphics[height=0.2\textwidth]{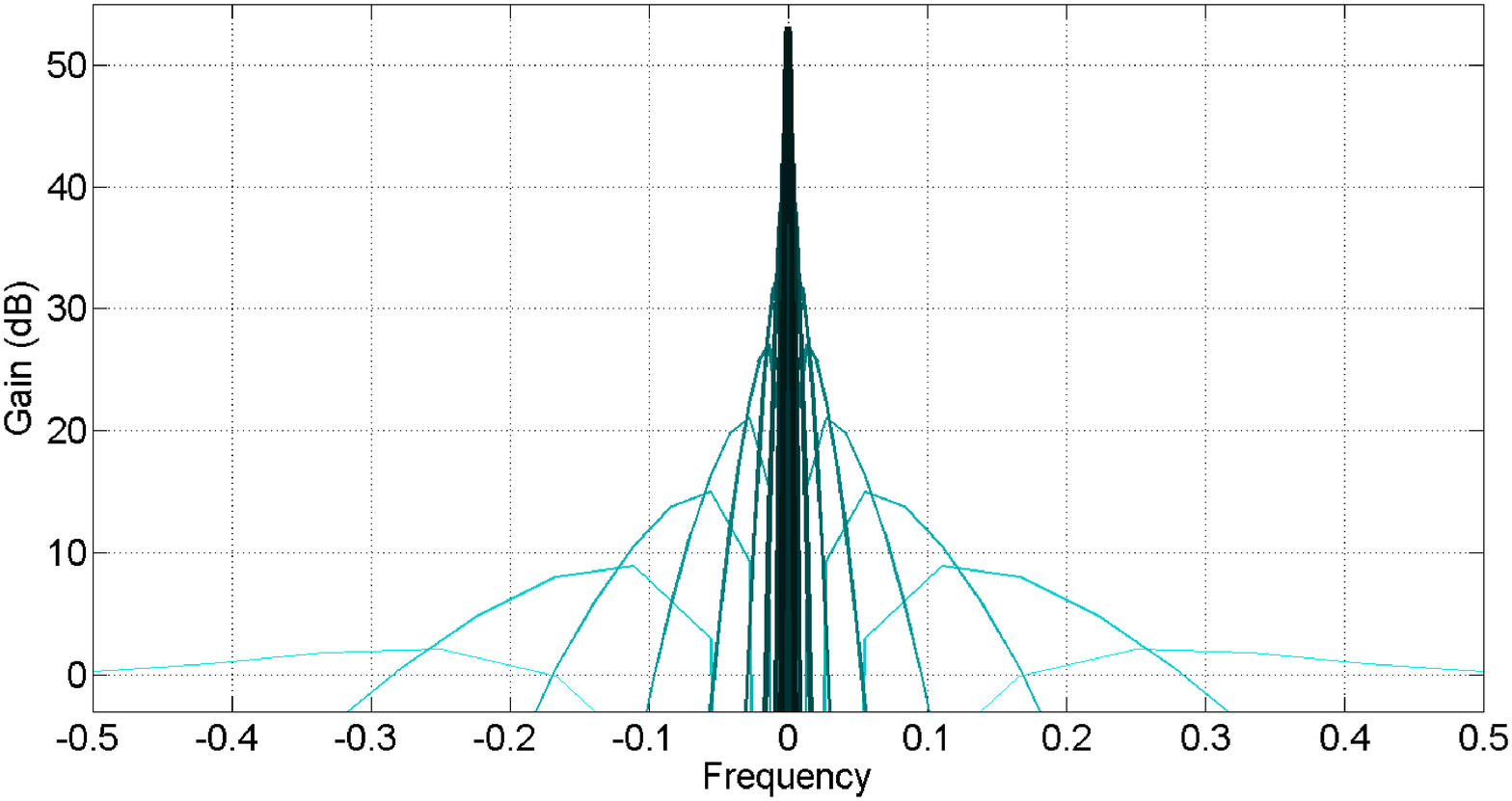} }
\subfigure[\label{fig:DemiSpectreSemiLogXY}]{\includegraphics[height=0.2\textwidth]{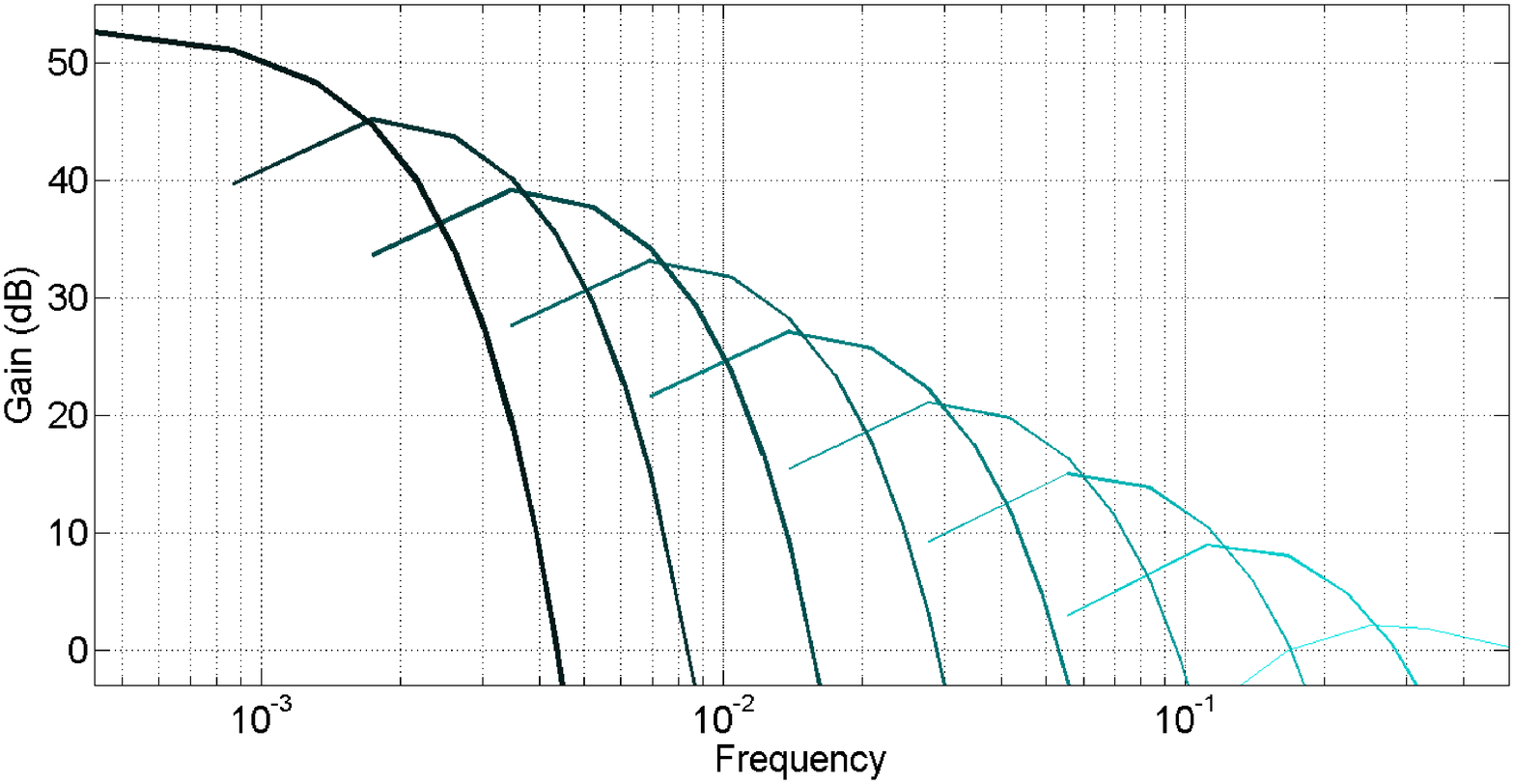} }
\subfigure[\label{fig:operateurTransformee}]{
\parbox[b]{0.2\textwidth}{
\begin{center}
$\longleftarrow\Phi\longrightarrow$\\[1mm]\includegraphics[height=0.175\textwidth]{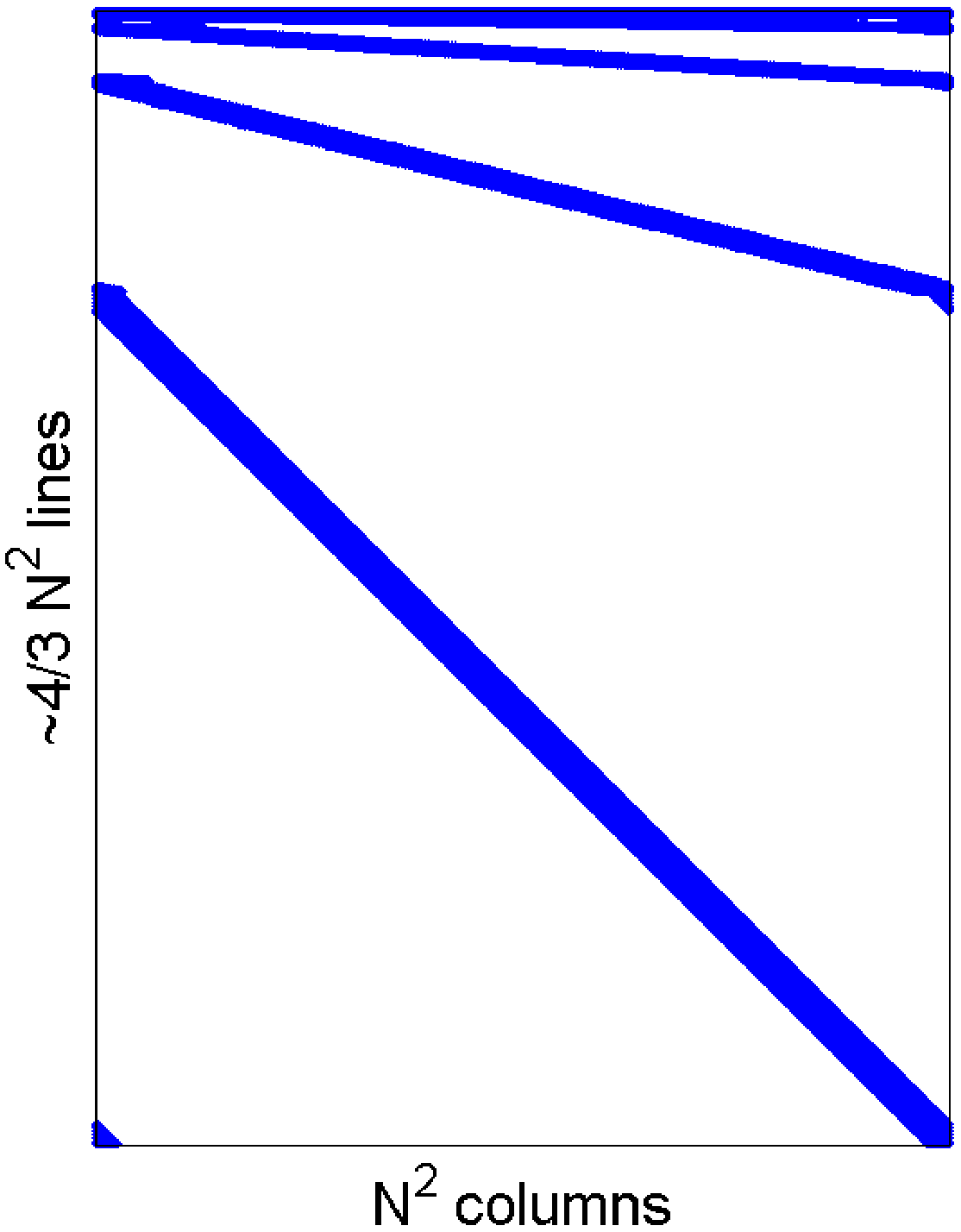}
\end{center}
}
}
}
\caption{~\ref{fig:spectreSemiLogY}: Spectrum of the DoG filters. The abscissa represents the frequencies.
The ordinate axis represents the different $DoG_k$ filters gain in dB.~\ref{fig:DemiSpectreSemiLogXY}: Half of the spectrum in~\ref{fig:spectreSemiLogY} with the abscissa having a logarithmic step. 
The scaling function $DoG_0$ is plotted in black thick line.
~\ref{fig:operateurTransformee}: In this paper, the transform $\Phi$ is represented as a matrix where blue dots correspond to a non-zero element. Note here that $\Phi$ is a highly sparse matrix.
}
\end{figure}

Indeed, the Fourier transform of a Gaussian is another Gaussian, so that $\mathscr{F}(DoG_{k})$ is a difference of Gaussians. We can easily prove, for $w^c=w^s=1$ (cf.~\eqref{eq:dogk}), that the central Fourier coefficient $\mathscr{F}(DoG_{k})\big(u_0(0), u_0(0)\big)=0$, and that $\mathscr{F}(DoG_{k})(i, j)>0$, $\forall (i,j)\neq \big(u_0(0), u_0(0)\big)$. This assertion can be verified on Figure~\ref{fig:DemiSpectreSemiLogXY}.
So, we replaced the $DoG_0$ filter, with no change in the notation, by a Gaussian low-pass scaling function consisting in its central component, with:
\begin{equation}
DoG_0 (x, y) = w^c G_{{\sigma}^c_0}(x,y). 
\end{equation} 
We do this to cover up the centre of the spectrum which otherwise won't be mapped. Figures~\ref{fig:spectreSemiLogY} and~\ref{fig:DemiSpectreSemiLogXY} show the spectrum partitioning with the DoG filters and the plug added by the scaling function $DoG_0$ (in black thick line). 
With no scaling function, all constant images would be mapped into the null image $0$ and this would make the transform be non-invertible. Here we overcome this problem as the central Fourier coefficient $\mathscr{F}(DoG_{0})\big(u_0(0), u_0(0)\big)>0$.

The scaling function introduction is further justified by the actual retina behavior. Indeed, the surround $G_{\sigma^s_k}$ in~\eqref{eq:dogk} appears progressively across time driving the filter passband from low frequencies to higher ones. So that, the Gaussian scaling function represents the very early state of the retina cells.

Once we introduced the $DoG_0$ scaling function, and in order to define an inverse for the new transform we must demonstrate that it is a ``frame''.  This will be detailed in Sections~\ref{sec:ecritureMatricielle} and~\ref{sec:preuveFrame}.

\subsection{Analysis and synthesis of an image using matrices}
\label{sec:ecritureMatricielle}
Unlike the implementations in~\cite{VanRullen01,Perrinet04,Masmoudi2010}, we use a matrix $\Phi$ to compute the image  transform through the modelled retina. The lines of $\Phi$ are the different analyzing $DoG_k$ filters. 
This yields an ``undersampled Toeplitz-block'' sparse matrix (see Figure~\ref{fig:operateurTransformee}).
Such an implementation allow fast computation of the multi-scale retinal transform through sparse matrix specific algorithms. 
This will in addition help us construct the dual frame of $\Phi$. The DoG transform is outlined in the following equation:
\begin{equation}
c = \Phi \, f.
\end{equation}
Interestingly, the straightforward synthesis as defined in~\eqref{eq:definitionReconstruction} amounts to the multiplication of the vector output $c$ by $\Phi^*$ the Hermitian transpose of $\Phi$. 
The reconstruction procedure is then outlined in the following equation:
\begin{equation}
\tilde{f}_{N_s} = \Phi^* \, c.
\end{equation}
\subsection{The DoG transform is a frame operator}
\label{sec:preuveFrame}
Our aim is to prove that the bio-inspired retina image transform presented amounts to a projection of the input image $f$ onto a frame of a vector space. The frame is a generalization of the idea of a basis to sets which may be linearly dependent~\cite{
Kovacevic2008}. These frames allow a redundant signal representation which, for instance, can be employed for coding with error resilience.  By proving the frame nature of this transform, we will be able to achieve an exact reverse transform through the construction of a dual frame. 
According to~\cite{
Kovacevic2008}, to prove that our transform is a frame one needs to show that $\exists\;0 < \alpha \leqslant \beta < {\infty}$,  such that:
\begin{equation} 
\label{eq:frameCondition} 
\alpha\,{\Vert f \Vert}^2 \leqslant \sum_{\kij \in \Gamma} \left(\ioplb_{\kij}\right)^2 \leqslant \beta\,{\Vert f \Vert}^2,\;\forall f.
\end{equation}
For example, authors in~\cite{rakshit1995error} proved that the Laplacian pyramid~\cite{Burt1983} is a frame and estimated experimentally $\alpha$ and $\beta$. The authors in~\cite{do2003framing} also designed an invertible Laplacian pyramid based on the same principles. Though there are some similarities, the DoG transform do not use the same filters and do not work in the same fashion. Thus, in the following, we give an original demonstration in our specific case which proves that the DoG transform is a frame, this furthermore with a mathematical formalism.

Let us prove these two inequalities of the so-called ``frame condition''.
\begin{proof}
\textbf{\textit{Upper bounding:}} 
We have:
\begin{equation}
\label{eq:developpementEnergieSousBandes}
\sum_{\kij \in \Gamma} \left(\ioplb_\kij\right)^2 = {\sum}_{k=0}^{K-1} \Vert B_k \Vert^2 ,
\end{equation}
where $B_k$ is the subband of scale $k$ generated by the image transform with:
\begin{align}
\label{eq:developpementEnergieGrille}
B_k(i,j) &=& \sum_{x=u_k(i)-M_k}^{u_k(i)+M_k}\sum_{y=u_k(j)-M_k}^{u_k(j)+M_k}DoG_{k}(u_k(i)-x, u_k(j)-y) \, f(x, y).\nonumber
\end{align}
If we denote by $U_k$ the undersampling operator corresponding to the function $u_k$ (cf. Equation~\eqref{eq:uk}), we can write the following:
\begin{equation}
B_k = U_k (DoG_k * f).
\end{equation}
Then, we have the following obvious inequalities:
\begin{align}
\Vert B_k \Vert = \Vert U_k (DoG_k * f)\Vert 
&\leqslant \left\Vert U_k \left(\vert DoG_k \vert * \vert f \vert\right)\right\Vert \leqslant \Vert \vert DoG_k \vert * \vert f \vert \Vert 
\leqslant \Vert DoG_k \Vert \, \Vert f \Vert .&\nonumber
\end{align}
So, with~\eqref{eq:developpementEnergieSousBandes} we infer the following bounding:
\begin{align}
\sum_{\kij \in \Gamma} \left(\ioplb_\kij\right)^2 = \sum^{K-1}_{k=0}\Vert B_k \Vert^2 & \leqslant \left(\sum^{K-1}_{k=0}\;{\Vert DoG_k \Vert}^2 \right){\Vert f \Vert}^2 = \beta \Vert f \Vert^2, &
\end{align}
which shows the first inequality.
\\
\textbf{\textit{Lower bounding:}} 
We start from the fact that:
\begin{equation}
\sum^{K-1}_{k=0}{\Vert B_k \Vert}^2 \geqslant {\Vert B_{K-1} \Vert}^2+{\Vert B_{0} \Vert}^2,
\end{equation}
which amounts to write the following inequalities:
\begin{align}
\label{eq:minorationIntermediare}
\sum_{\kij \in \Gamma}\;\left( \ioplb_\kij \right)^2 & \geqslant 
{\Vert DoG_{K-1} * f \Vert}^2 + {\Vert \big(DoG_{0} * f\big) \big(u_0(0), u_0(0)\big)\Vert}^2 &\nonumber\\
& =  {\Vert \mathscr{F}(DoG_{K-1}) \, \mathscr{F}(f) \Vert}^2  + {\Vert \big(\mathscr{F}(DoG_{0}) \, \mathscr{F}(f)\big) \big(u_0(0), u_0(0)\big)\Vert}^2 &\nonumber\\
 & = \sum^{N-1}_{i,j = 0}{\big( \mathscr{F}(DoG_{K-1})(i,j) \, \mathscr{F}(f)(i,j) \big)}^2 +
{\Vert \mathscr{F}(DoG_{0})\big(u_0(0), u_0(0)\big) \, \mathscr{F}(f)\big(u_0(0), u_0(0)\big)\Vert}^2,  &\nonumber
\end{align}
We know that $DoG_{K-1}(i, j)>0, \; \forall (i,j)\neq \big(u_0(0), u_0(0)\big)$ and that $DoG_{K-1}\big(u_0(0), u_0(0)\big)=0$. We also have $\mathscr{F}(DoG_{0})\big(u_0(0), u_0(0)\big)>0$ (cf. Section~\ref{sec:scalingFunction}). 
So, if we define a real $\alpha$ by:
\begin{equation}
\label{eq:definitionB}
\alpha = min\Big\lbrace\big\lbrace\mathscr{F}(DoG_{0})^2\big(u_0(0), u_0(0)\big)\big\rbrace \cup \big\lbrace\mathscr{F}(DoG_{K-1})^2(i,j),\, (i,j) \in {\llbracket 0, N-1\rrbracket}^2\setminus\big(u_0(0),u_0(0)\big) \big\rbrace\Big\rbrace ,\nonumber
\end{equation}
then we have that $\alpha>0$, and we get the following inequalities:
\begin{align*}
& \sum^{N-1}_{i,j = 0}
{\big( \mathscr{F}(DoG_{K-1})(i,j) \, \mathscr{F}(f)(i,j) \big)}^2 + {\Vert \mathscr{F}(DoG_{0})\big(u_0(0), u_0(0)\big) \, \mathscr{F}(f)\big(u_0(0), u_0(0)\big)\Vert}^2  &\\
= & \hspace*{-29mm}\sum_{\hspace*{29mm}i,j \in {\llbracket 0, N-1\rrbracket}^2\setminus(u_0(0),u_0(0))}\hspace{-29mm}{\big( \mathscr{F}(DoG_{K-1})(i,j) \, \mathscr{F}(f)(i,j) \big)}^2 + {\Vert \mathscr{F}(DoG_{0})\big(u_0(0), u_0(0)\big) \, \mathscr{F}(f)\big(u_0(0), u_0(0)\big)\Vert}^2  & \\
\geqslant & \;\alpha\,\hspace*{-11mm}\sum_{\hspace*{11mm}i,j \in {\llbracket 0, N-1\rrbracket}^2}\hspace*{-12mm}{\big( \mathscr{F}(f)(i,j) \big)}^2 =  \alpha\Vert f\Vert^2, & 
\end{align*}
so that, $\sum_{\kij \in \Gamma} \left(\ioplb_{\kij}\right)^2 \geqslant \alpha\Vert f\Vert^2$ which concludes the proof.
\end{proof}
\subsection{Synthesis using the dual DoG frame}
\label{sec:frameDuale}

We introduce in this section a correction means for the reconstruction error in the retina model presented through the frame theory.

\begin{SCfigure}
\begin{tabular}{cc}
{\includegraphics[width=0.3\textwidth]{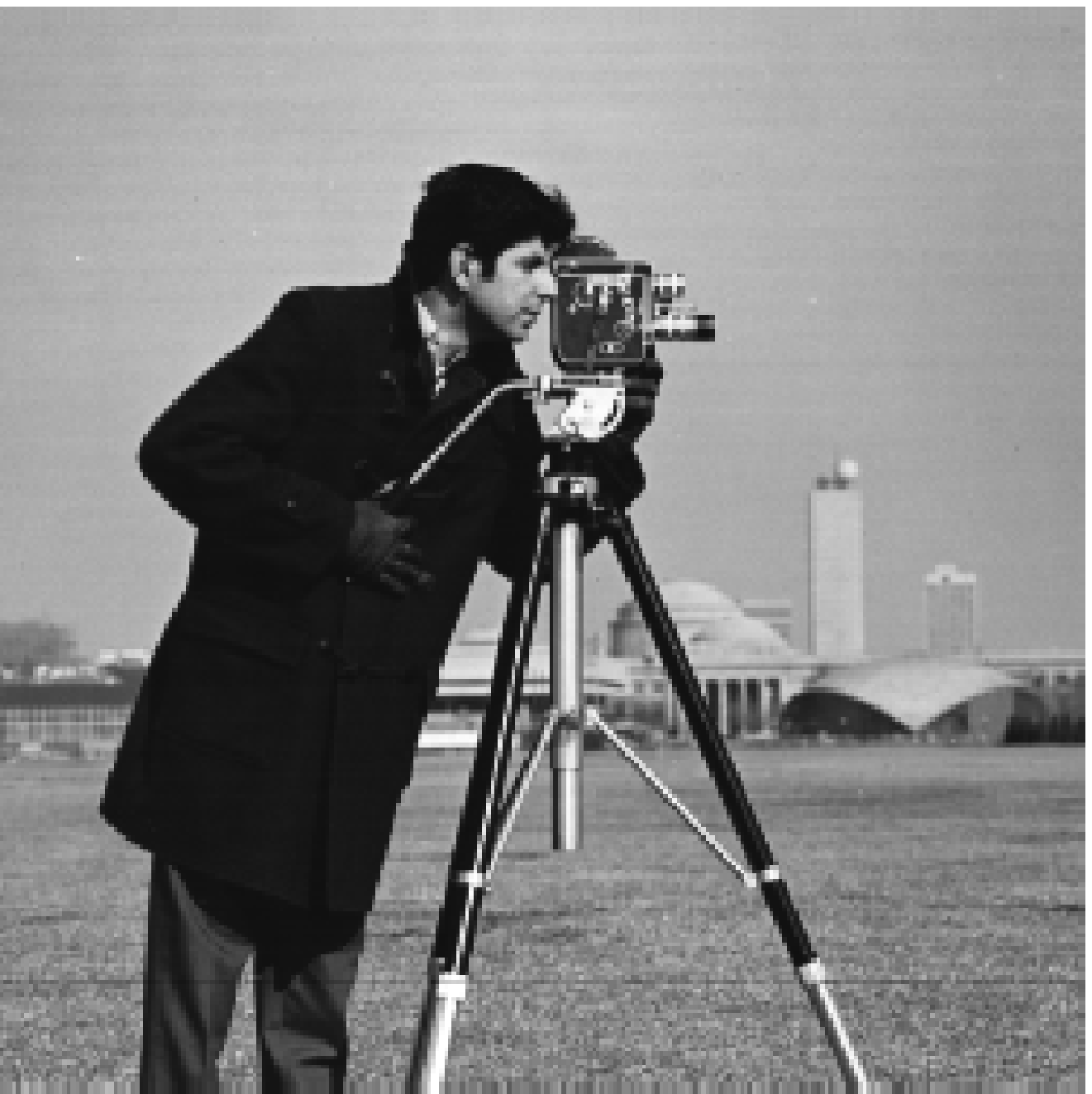} }&
{\includegraphics[width=0.3\textwidth]{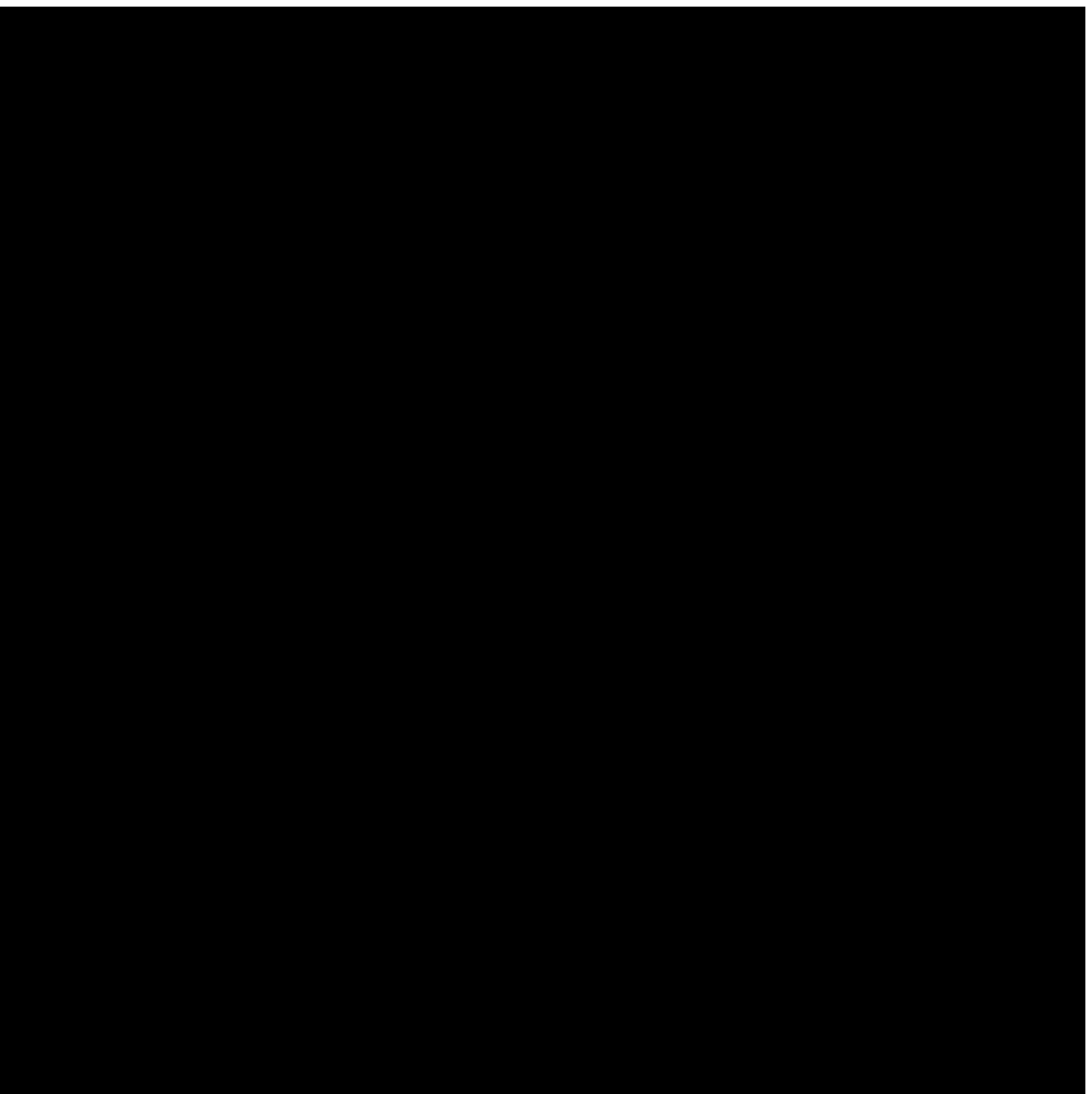} }\\
(a) & (b)
\end{tabular}
\caption{\label{fig:resultats}
Result of the decoding procedure with the dual DoG frame using all of the retina cells responses.~\ref{fig:resultats}(a): Reconstructed image. The PSNR quality measure of ${f}^{*}_{N_s}$ yields $296$ dB.~\ref{fig:resultats}(b): Error image in logarithmic scale. 
This shows that 
the reconstruction using the dual frame is very precise.
}
\end{SCfigure}

The straightforward analysis/synthesis procedure can be outlined in the relation between the input image and the reconstruction estimate:
$\tilde{f}_{N_s} = \Phi^* \Phi f.$ 
As we already demonstrated that the DoG transform is a frame, $\Phi^* \Phi$ is said to be the frame operator. To have an exact reconstruction of $f$, one must construct the dual DoG vectors. A preliminary step is to compute $(\Phi^*\Phi)^{-1}$, the inverse frame operator. We then get a corrected reconstruction $f^{*}_{N_s}$, defined by: $f^*_{N_s} = (\Phi^*\Phi)^{-1}\tilde{f}_{N_s}$. If $N_s$ is the total number of the retina model cells, we have:
\begin{equation}
\label{eq:reconstructionExacte}
f^*_{N_s} = (\Phi^*\Phi)^{-1}\tilde{f}_{N_s} =
\left((\Phi^*\Phi)^{-1}\Phi^*\right)\;\ioplb = 
\left((\Phi^*\Phi)^{-1}\Phi^*\right)\;\Phi \, f = 
f.
\end{equation}
As made clear through Equation~\eqref{eq:reconstructionExacte}, the dual vectors are the lines of $(\Phi^*\Phi)^{-1}\Phi^*$. If we reconstruct $f^*_{N_s}$ starting from the ROC output $c$ and using the dual frame vectors, we get the results shown in Figure~\ref{fig:resultats}. The reconstruction obtained is accurate and requires only a simple matrix multiplication. In this example, reconstruction quality is evaluated to $296$ dB of PSNR. \\

$(\Phi^*\Phi)^{-1}$ is a square, definite positive invertible matrix~\cite{Kovacevic2008}. Another advantage of our method is that $(\Phi^*\Phi)^{-1}$ is well conditioned, with a conditioning number estimated to around $16$, so that its inversion is stable. This is a crucial issue as previous works aimed at conceiving the DoG reverse transform tried to invert the original filter bank with no scaling function $DoG_0$~\cite{Sen07,Bhattacharya2010}. This is obviously mathematically incorrect as the filter bank thus defined is not a frame and thus its pseudo inverse $(\Phi^*\Phi)^{-1}\Phi^*$ does not exist. The solution proposed by the authors of~\cite{Sen07,Bhattacharya2010} gives only a least squares solution to an ill-conditioned problem. Our method instead is stable. Besides through the out-of-core algorithm that we designed we can invert $(\Phi^*\Phi)$ even for large images whereas authors in~\cite{Sen07,Bhattacharya2010} are restricted to a maximum size of $32 \times 32$.

Furthermore, correcting the reconstruction errors using the adequate dual frame does not alter the coding procedure. Indeed, methods introduced in~\cite{Perrinet04,Bhattacharya2010} require a  Matching Pursuit (MP) procedure. MP is time consuming and depends on the order in which the ``match and update'' mechanism is performed. Our method keeps the coding procedure straightforward, fast and order-independent.

\paragraph*{The recursive out-of-core blockwise inversion algorithm:}
Though the mathematical fundamentals underlying this work are simple, the implementation of such a process is a hard problem. Indeed, in spite of the sparsity of $\Phi$ and $\Phi^*$, the frame operator $\Phi^*\Phi$ is an $N^4$-sized dense matrix for an $N^2$-sized image $f$. For instance, if $N=257$, $\Phi^*\Phi$ holds in $16$ Gbytes, and $258$ Gbytes if $N=513$.
The solution is to recourse to the so-called out-of-core (OOC) algorithms~\cite{toledo1999survey}. 

The frame operator $\Phi^*\Phi$ is constructed block by block, and each block is stored separately on disk. The inversion is then performed using a recursive algorithm that relies on the block-wise matrix inversion formula that follows:
\begin{equation}
\left(
\begin{array}{cc}
A & B\\ 
C & D
\end{array} 
\right)^{-1} =
\left(
\begin{array}{ccc}
A^{-1}+A^{-1}\,B\,Q^{-1}\,C\,A^{-1} &\;\;& -A^{-1}\,B\,Q^{-1}\\ 
-Q^{-1}\,C\,A^{-1} &\;& Q^{-1}
\end{array} 
\right),
\end{equation}
where $Q$ is the Schur complement of A, such that:
$Q = D-C\,A^{-1}\,B.$
Thus, inverting a matrix amounts to the inversion of two matrices that are 4 times smaller. The inversion consists then in subdividing the problem by a factor 4 at each recursion level until we reach a single block problem. Obviously, this algorithm requires OOC blockwise matrix routines for multiplication, subtraction and addition, that we implemented in parallel to accelerate the computation. 

\begin{figure}[h!]
\begin{tabular}{p{7.5mm}m{0.18\textwidth}m{0.18\textwidth}m{0.18\textwidth}m{0.18\textwidth}m{0.18\textwidth}}
{\large $\tilde{f}_{N_s}$}
&
\includegraphics[width=0.18\columnwidth]{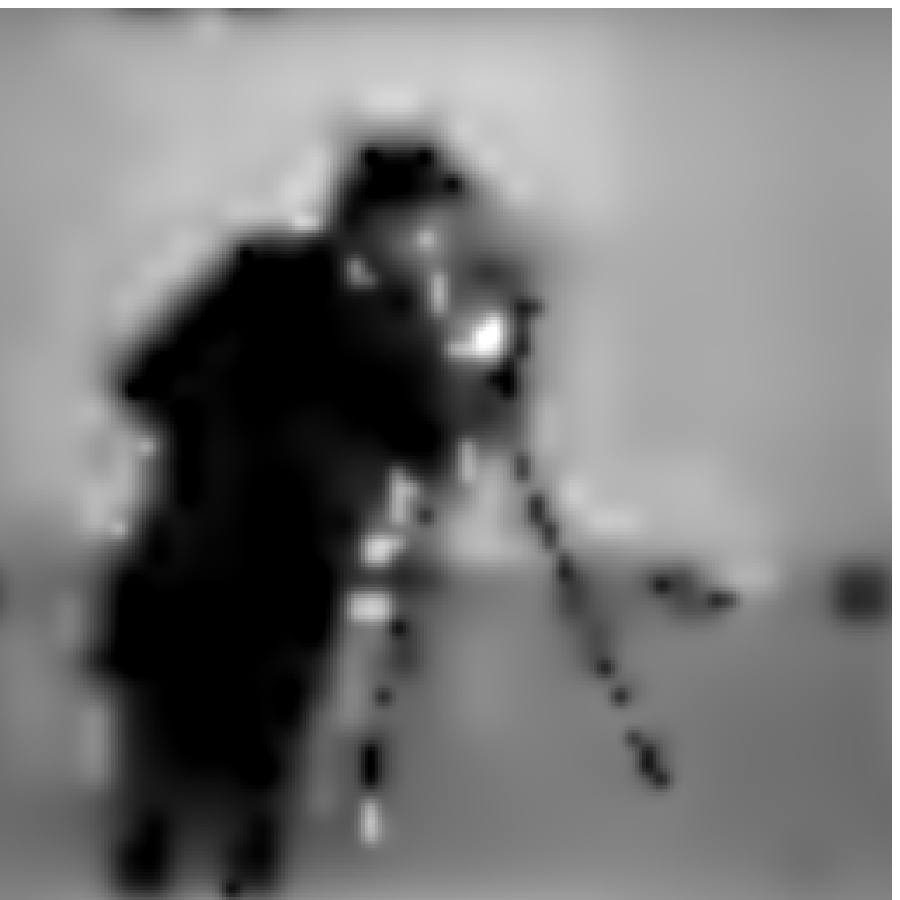} &
\includegraphics[width=0.18\columnwidth]{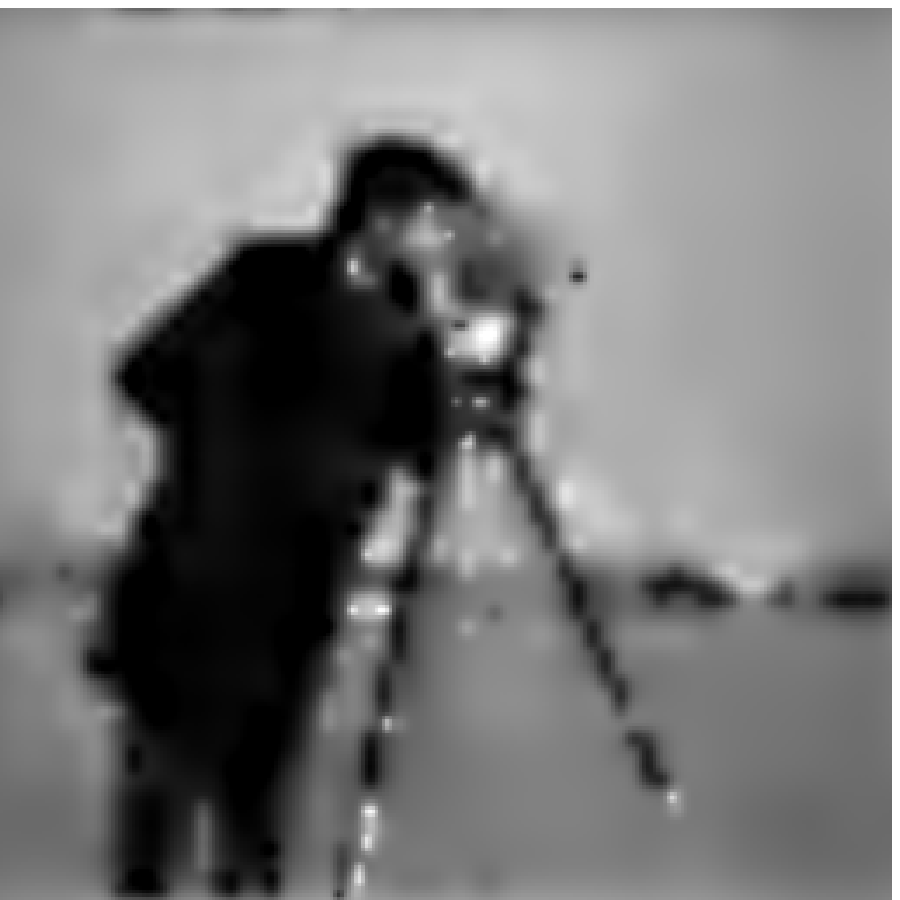}   &
\includegraphics[width=0.18\columnwidth]{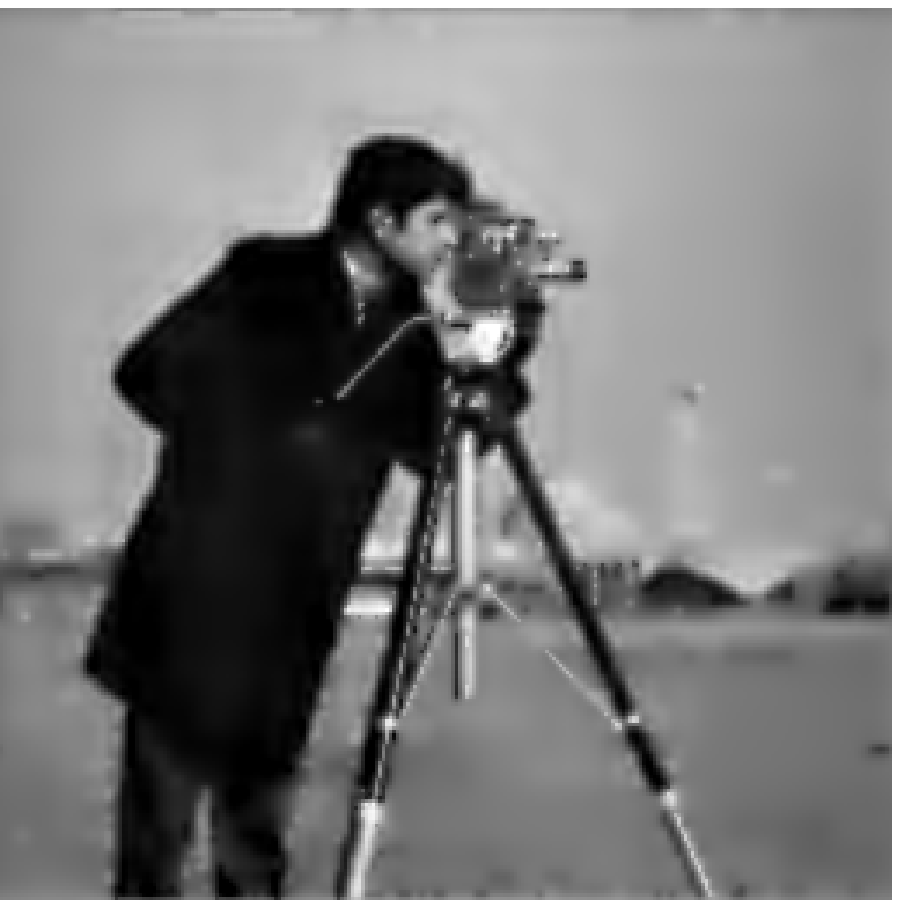}   &
\includegraphics[width=0.18\columnwidth]{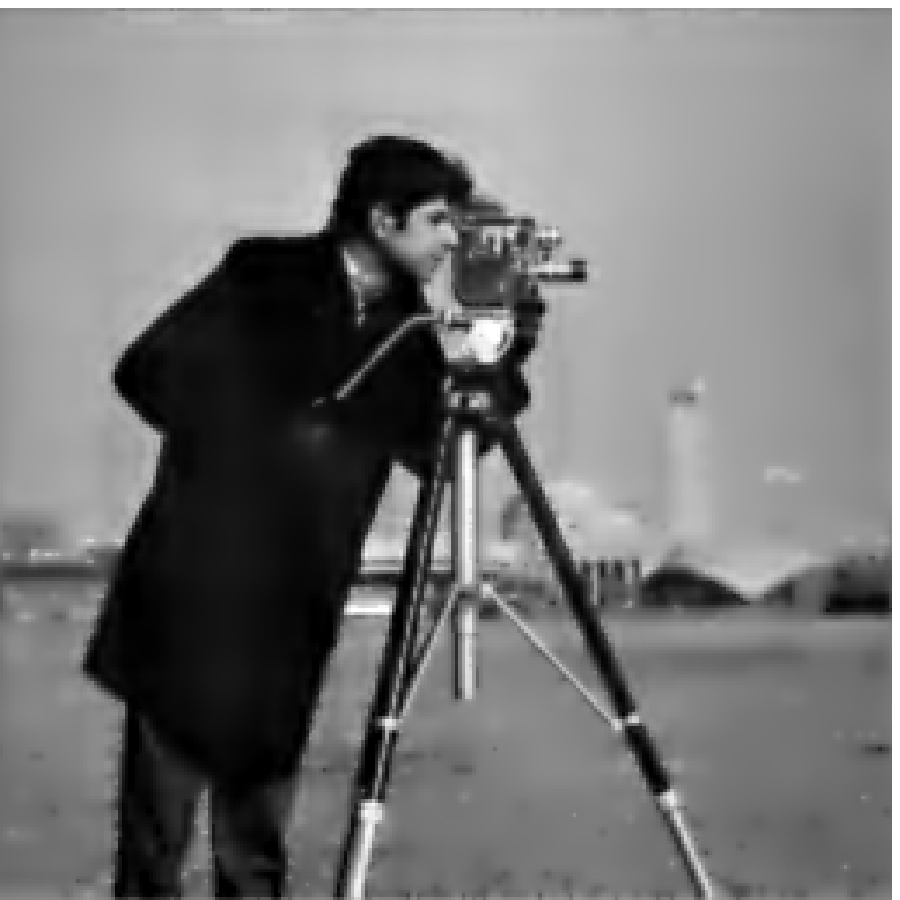}  &
\includegraphics[width=0.18\columnwidth]{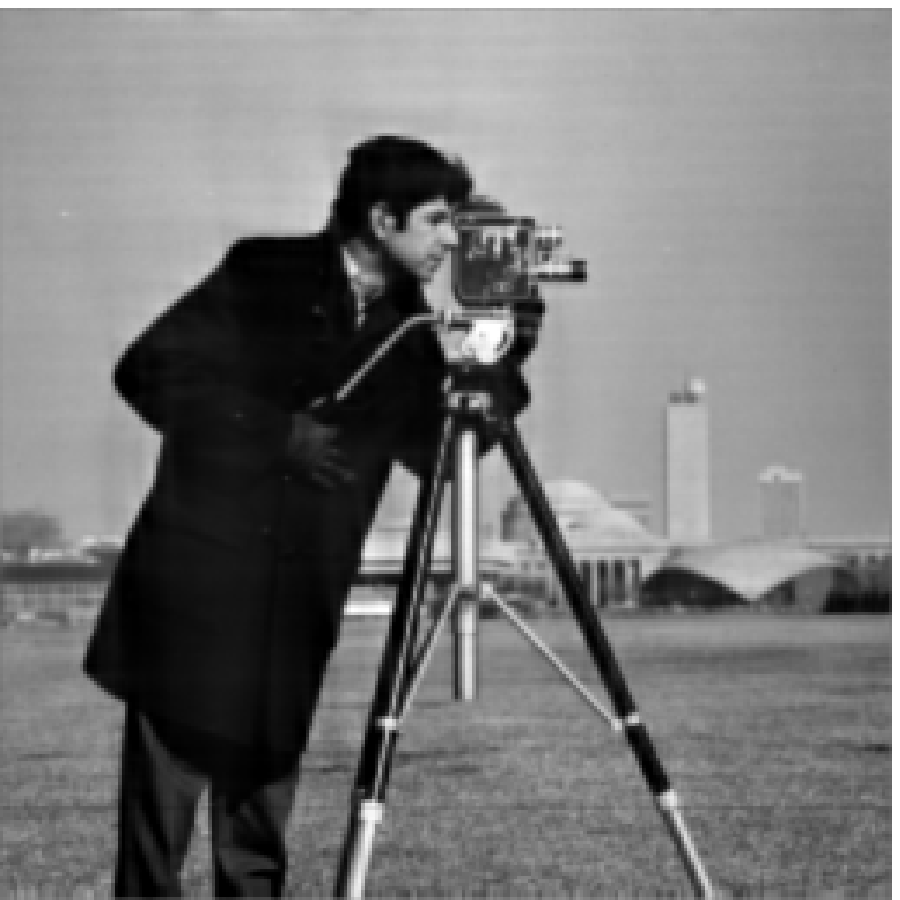}  \\
{\large $f^{*}_{N_s}$}
&
\includegraphics[width=0.18\textwidth]{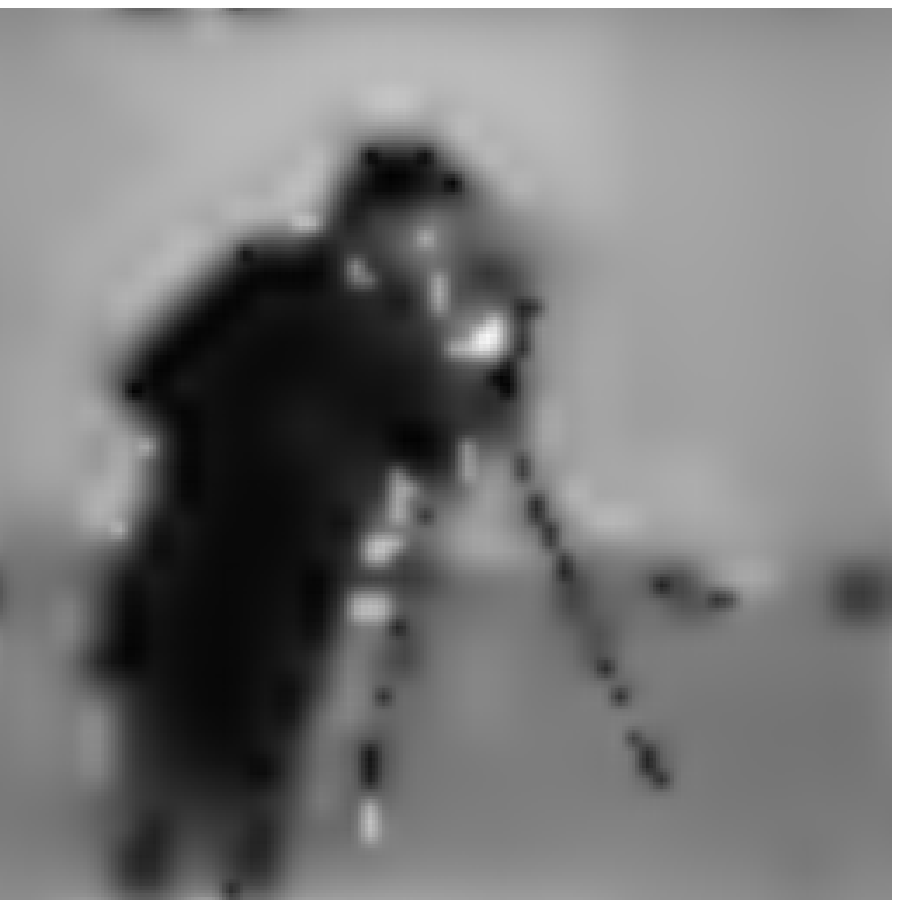} &
\includegraphics[width=0.18\textwidth]{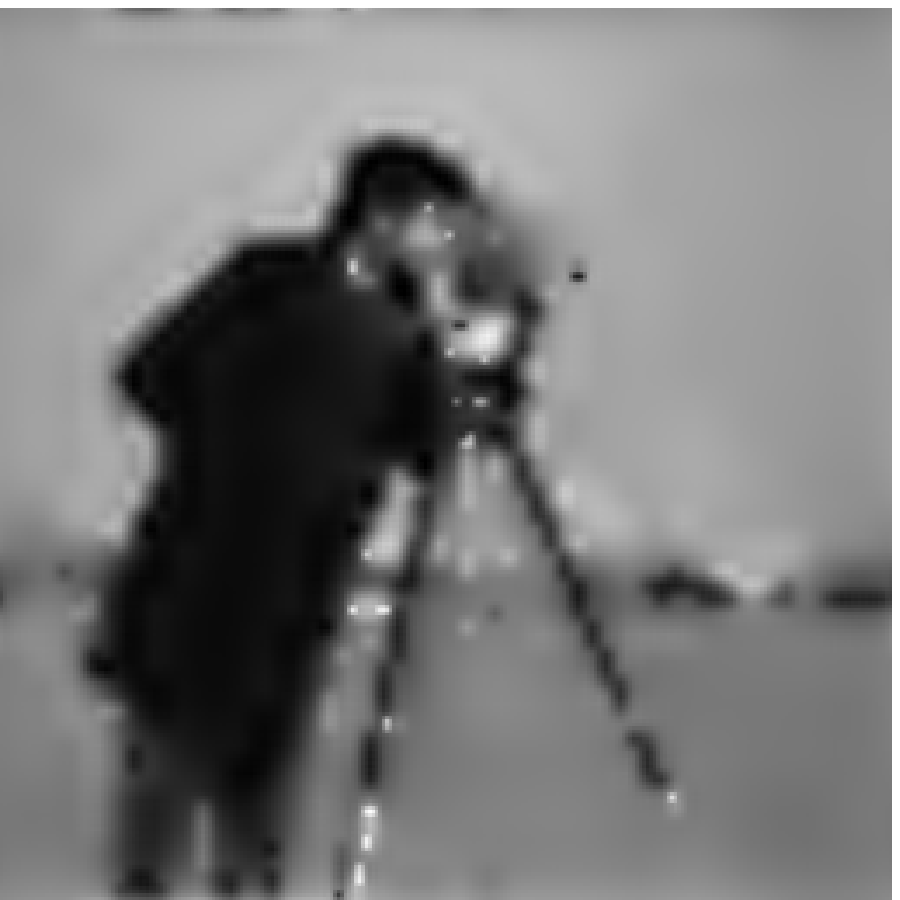}   &
\includegraphics[width=0.18\textwidth]{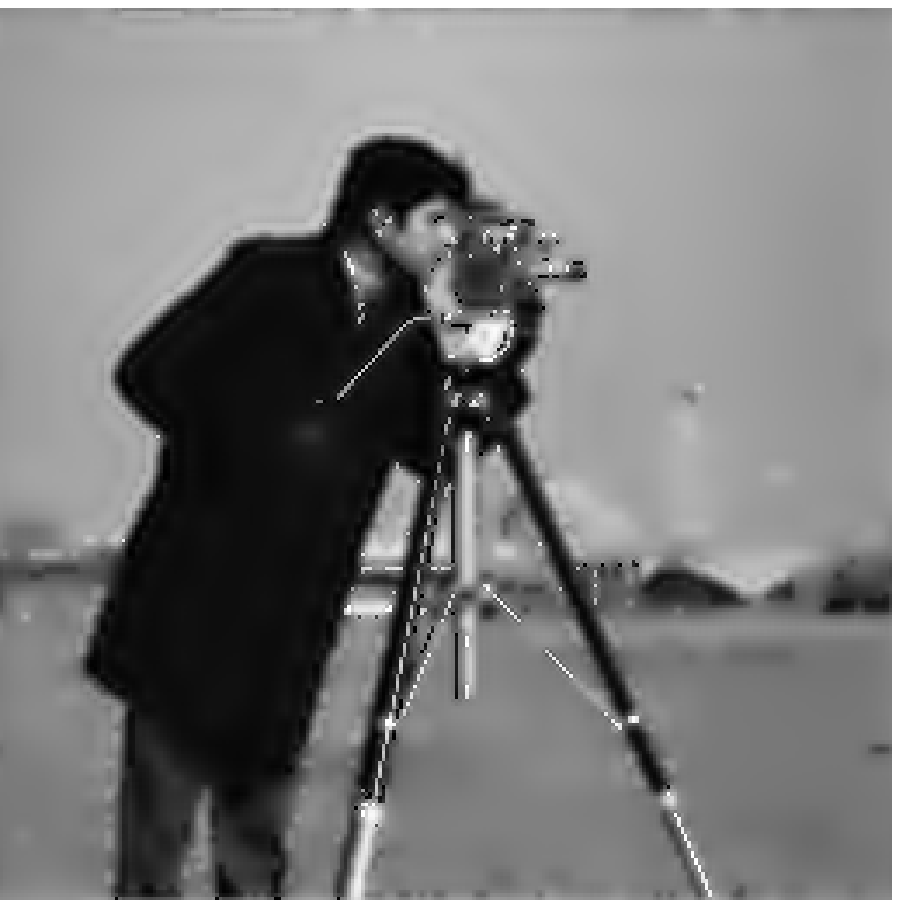}   &
\includegraphics[width=0.18\textwidth]{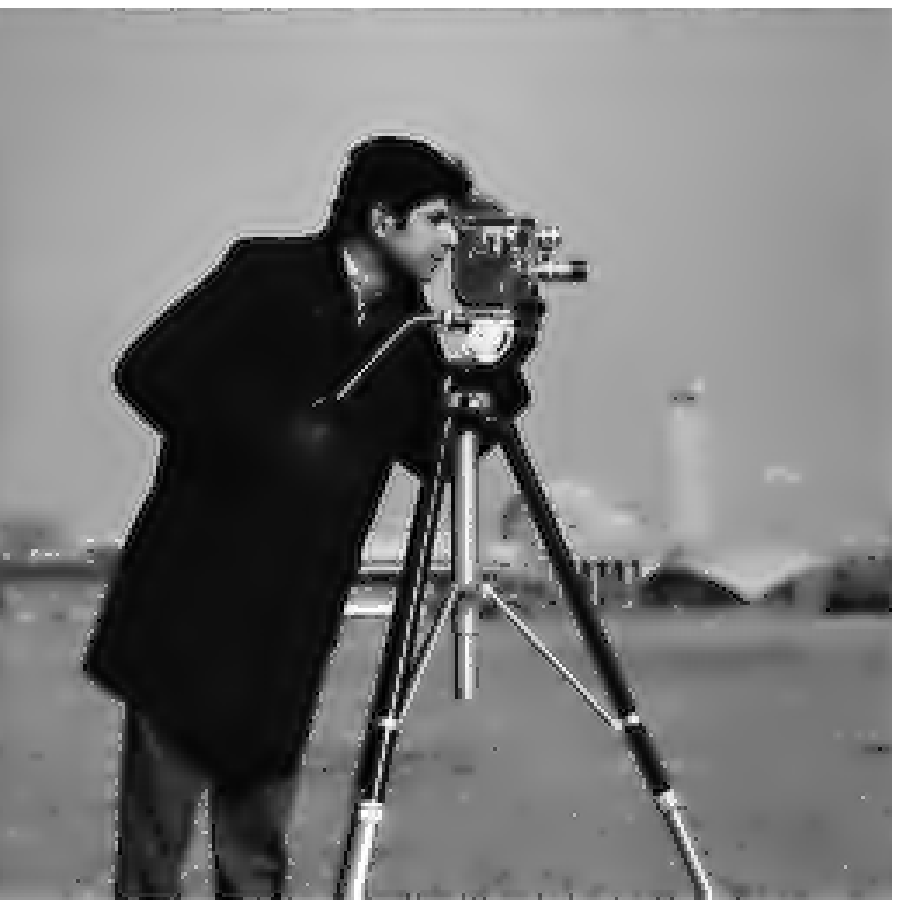}  &
\includegraphics[width=0.18\textwidth]{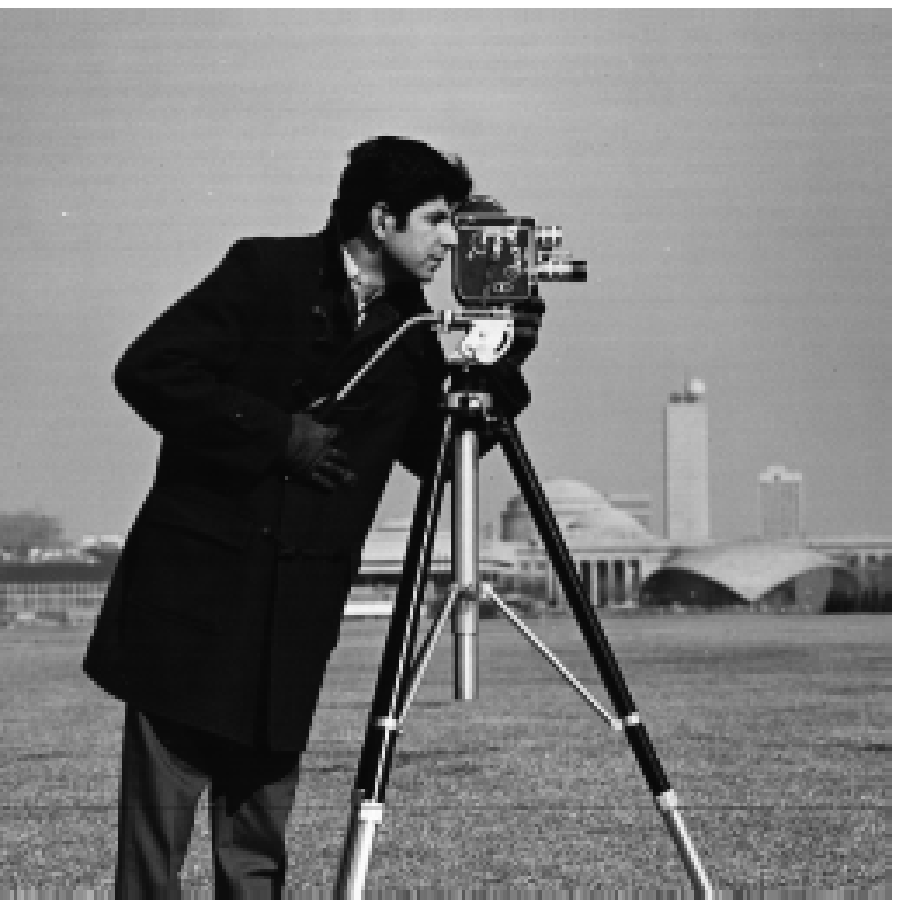}  \\
&
\begin{center}
$0.5\%$
\end{center} 
&
\begin{center}
$1\%$ 
\end{center}
&
\begin{center}
$5\%$ 
\end{center}
&
\begin{center}
$10\%$ 
\end{center}
&
\begin{center}
$100\%$
\end{center}
\end{tabular}
\caption{\label{fig:reconstructionProgressiveROC}Reconstruction of the cameraman image $f$ using different percentages of the highest $\ioplb_\kij$. 
PSNR for the upper/lower image is from left to right: (19.2 dB/19.5 dB), (20.4 dB/20.8 dB), (24.08 dB/25 dB), (25.8 dB /27.5 dB), and (27.9 dB/296 dB).}
\end{figure}

\section{Comparison to the original rank order codec}
\label{sec:results}
We experiment our new decoder in the context of scalable image decoding. We reconstruct the \textit{cameraman} test image using an increasing number $N_s$ of significant coefficients (cf. Equation~\eqref{eq:definitionReconstruction}). We compare the results when using the original DoG filters in $\Phi$ and the dual DoG filters $(\Phi^{*}\Phi)^{-1}\Phi^{*}$ for the decoding procedure. 
Figure~\ref{fig:reconstructionProgressiveROC} summarizes the results obtained, with the upper line showing the progressive straightforward reconstruction $\tilde{f}_{N_s}$ and the bottom line showing the corrected progressive reconstruction $f^{*}_{N_s}$. The gain in PSNR is significant for low rates (around 0.3 dB) and increases with $N_s$ up to 270 dB.
Besides, our method does not alter the coding procedure and keeps it straightforward, fast and order-independent.

\bibliographystyle{splncs03}
\bibliography{IEEEabrv,bibliography}

\end{document}